\documentclass[letterpaper, 10 pt, conference]{ieeeconf}  

\IEEEoverridecommandlockouts                              

\usepackage[utf8]{inputenc}

\title{\LARGE Genie: Smart ROS-based Caching for Connected Autonomous Robots}
 \author{
 {Zexin Li$^{1*}$, Soroush Bateni$^{2*}$, Cong Liu$^{1}$}\\
\thanks{$^{1}$Authors are with the University of California, Riverside.}
\thanks{$^{2}$Authors are with the University of Texas at Dallas.}
\thanks{$^{*}$Authors make equal contributions.}
 } 

\usepackage{cite}
\usepackage{caption} 
\usepackage{subfig}
\usepackage{graphicx}
\usepackage{textcomp}
\usepackage{amsmath,amssymb,amsfonts}

\usepackage{amsthm}
\usepackage{algorithm}
\usepackage{algorithmicx}
\usepackage[noend]{algpseudocode}
\usepackage{listings}
\usepackage{xcolor}

\usepackage[hyphens]{url}
\usepackage[hidelinks]{hyperref}
 
\usepackage{cleveref}
\usepackage[makeroom]{cancel}
\usepackage{xcolor,colortbl}
\def\BibTeX{{\rm B\kern-.05em{\sc i\kern-.025em b}\kern-.08em
    T\kern-.1667em\lower.7ex\hbox{E}\kern-.125emX}}

\usepackage{tikz}
\usetikzlibrary{positioning}
\usepackage{multirow,booktabs,multicol}

\usepackage{url}

\newcommand\hmm[1]{\ifnum\ifhmode\spacefactor\else2000\fi>1000 \uppercase{#1}\else#1\fi}

\usepackage{setspace}
\usepackage{xspace}
\usepackage[export]{adjustbox}

\let\labelindent\relax
\usepackage{enumitem}

\algnewcommand{\algorithmicgoto}{\textbf{go to}}%
\algnewcommand{\Goto}[1]{\algorithmicgoto~\ref{#1}}%

\makeatletter
\newcommand*{\rom}[1]{\expandafter\@slowromancap\romannumeral #1@}
\makeatother

\usepackage{multirow,booktabs,multicol}
\begin{document}

\maketitle

\thispagestyle{plain}
\pagestyle{plain}


\begin{abstract}
Despite the promising future of autonomous robots, several key issues currently remain that can lead to compromised performance and safety. One such issue is latency, where we find that even the latest embedded platforms from NVIDIA fail to execute intelligence tasks (e.g., object detection) of autonomous vehicles in a real-time fashion. One remedy to this problem is the promising paradigm of edge computing. Through collaboration with our industry partner, we identify key prohibitive limitations of the current edge mindset: (1) servers are not distributed enough and thus, are not close enough to vehicles, (2) current proposed edge solutions do not provide substantially better performance and extra information specific to autonomous vehicles to warrant their cost to the user, and (3) the state-of-the-art solutions are not compatible with popular frameworks used in autonomous systems, particularly the Robot Operating System (ROS).

To remedy these issues, we provide Genie, an encapsulation technique that can enable transparent caching in ROS in a non-intrusive way (i.e., without modifying the source code), can build the cache in a distributed manner (in contrast to traditional central caching methods), and can construct a collective three-dimensional object map to provide substantially better latency (even on low-power edge servers) and higher quality data to all vehicles in a certain locality. We fully implement our design on state-of-the-art industry-adopted embedded and edge platforms, using the prominent autonomous driving software Autoware, and find that Genie can enhance the latency of Autoware Vision Detector by 82\% on average, enable object reusability 31\% of the time on average and as much as 67\% for the incoming requests, and boost the confidence in its object map considerably over time. 
\end{abstract}

\section{Introduction}

Autonomous robots are rapidly expanding from a research-oriented subject to real-world applications, with autonomous vehicles being a particular and promising representation~\cite{song2024bundledslam,song2024eta,hawkins_2018,gupta_2019}. Companies such as Waymo already started deploying fully autonomous taxi fleets in a limited number of areas~\cite{hawkins_2018,gupta_2019}. However, several outstanding questions remain when it comes to accuracy, latency, timing-predictability, and energy efficiency. 

The problem addressed in this paper is that of latency given the Size, Weight, and Power (SWaP) constraints of the computing platform used in autonomous vehicles. Specifically, existing low-power autonomous embedded platforms such as the latest NVIDIA Jetson Xavier cannot execute crucial intelligence tasks such as object detection and localization in a real-time fashion. To remedy this, some existing work suggests that various server clusters can be strategically located close to the autonomous vehicle to aid in computational tasks~\cite{mao2017survey}.

In collaboration with Fujitsu, a leader in edge computing and networking, this work addresses the pivotal challenges in integrating edge computing with autonomous vehicle technologies. The primary obstacle is the necessity for edge computing infrastructure, particularly remote server clusters, to be in close proximity to the vehicles to mitigate communication costs, rendering the use of conventional, energy-intensive data center equipment both impractical and costly. Consequently, there's a compelling need for affordable, low-power alternatives.

\noindent\textbf{Key Insight}: Utilizing edge servers with locality-aware caching can reduce latency significantly if results can be reused effectively, making it feasible to deliver enhanced performance and provide additional, valuable environmental information to autonomous vehicles.

Nonetheless, existing solutions (reviewed in detail in Sec.~\ref{sec:relatedwork}) lack three crucial design features that are required for practical deployment of a locality-aware caching method: (1) incompatibility with the Robot Operating System (ROS), (2) reliance on parameter servers, or a central cache, and (3) lack of specific usability for autonomous vehicles.

\noindent\textbf{Contributions}: To address these issues, we introduce Genie, a distributed ROS-based interface for object-oriented caching and data sharing, constructing a 3D object map of the edge server's surroundings: (1) ROS-based Transparent Caching: Genie intercepts ROS communications for caching without modifying software. (2) Distributed Cache Construction: An algorithm allows Genies to communicate, enhancing their local caches. (3) 3D Object Map: Identifies and caches useful real-world objects for autonomous vehicles, reducing redundant computation and providing extra information.


\noindent\textbf{Implementation and Evaluation}: We implemented Genie on a vehicular edge computing platform with Jetson TX2s and AGX Xaviers, using Autoware~\cite{kato2018autoware} as a representative ROS-based system. Our evaluation shows that Genie reduces latency by 82\% on average, with a peak improvement of 95\%. Object reusability is effective, with 31\% reusable objects on average and up to 67\%. A confidence score demonstrates improved data quality through multi-car information gathering.

\section{Background and Motivation}


Autonomous Vehicles (AVs) rely on a multi-stage computational process involving sensors like cameras and LiDAR to ensure safe driving decisions~\cite{gog2021pylot}. This pipeline, which includes perception, localization, detection, prediction, planning, and control, must meet stringent timing constraints to maintain safety. However, state-of-the-art deep learning models impose heavy computational demands, and even advanced edge devices struggle to keep pace, forcing trade-offs in data and algorithm selection.

Our evaluation of six cutting-edge object detection models \cite{YOLO,carion2020end} across various GPU-enabled devices (Nano, AGX, Orin) and an edge server with an A4500 GPU shows significant speedups. For instance, YOLOv8s and DETR-ResNet-101 achieve speedups of 5.02x and 10.37x on the A4500 compared to the Nano, illustrating that advanced GPU-equipped edge servers can substantially enhance computational efficiency by reusing results. Note that for some small models like YOLOv8s, even all evaluated devices provides acceptable latency, but this model has the lowest accuracy.

The advent of edge servers presents a significant opportunity to augment computational efficiency. By offloading tasks to edge servers, AVs can speed up data processing and avoid system failures, such as out-of-memory errors observed with DETR-ResNet-101-DC5 on less capable hardware.

This motivates the incorporation of edge servers or powerful embedded devices as distributed caches to help AVs meet stringent timing constraints, enhance safety-critical processes, and improve overall performance.

\begin{table}[!tbp]
\centering

\caption{Runtime/speedup of autonomous driving models on different devices (Nano, AGX, Orin) and edge server GPU (A4500). OOM indicates an out-of-memory error. The baseline for speedup is the slowest successful execution among devices.}
\resizebox{0.5\textwidth}{!}{

\begin{tabular}{lcccc}
\toprule
Model & Nano & AGX & Orin & A4500 \\ 
\midrule
YOLOv8s             & 27.60 / 1.00x & 21.20 / 1.30x & 13.73 / 2.01x  & 5.50 / 5.02x \\
YOLOv8m             & 60.90 / 1.00x & 48.45 / 1.26x & 17.50 / 3.48x  & 7.10 / 2.46x \\
YOLOv8l             & 92.00 / 1.00x & 85.50 / 1.08x & 26.20 / 3.51x  & 9.70 / 2.70x \\
DETR-ResNet-50      & 307.28 / 1.00x & 230.39 / 1.33x & 112.26 / 2.74x  & 30.91 / 9.94x \\
DETR-ResNet-101     & 422.51 / 1.00x & 336.96 / 1.25x & 145.92 / 2.90x  & 40.73 / 10.37x \\
DETR-ResNet-101-DC5  & OOM / N/A & 747.30 / 1.00x & 316.52 / 2.36x  & 86.13 / 8.68x \\
\bottomrule
\end{tabular}
}
\label{tab:case}
\end{table}

\section{Design}

\subsection{Design Considerations}\label{sec:des:cons}

\noindent\textbf{Design Goals.} First and foremost, we would like to provide a design that can improve latency substantially when using low-power edge servers. In the process of our design, we would like to create an architecture that can improve decision-making accuracy by providing smarter, as well as more reliable, information to autonomous vehicles to offer an incentive to connect to edge services that are going to be imminent. To make our design practical, we design and implement everything around the ROS framework~\cite{quigley2009ros}, which is the most prominent framework used to implement autonomous vehicles. However, similar techniques can also be applied to other frameworks as long as they follow a modular peer-to-peer communication approach. Finally, we note that in all of our design decisions, the features that are added will always remain optional since the car will carry all the necessary computing modules for full autonomy.

\begin{figure}[t]
    \centering
    \includegraphics[width=0.45\linewidth,trim={0 12.7cm 26.3cm 0},clip]{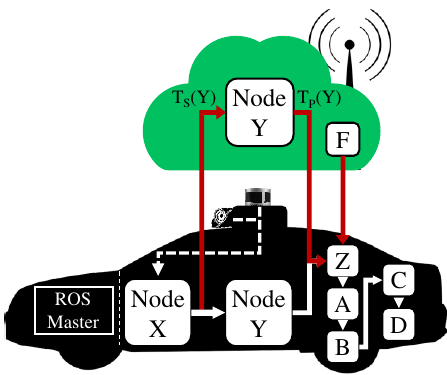}
    \includegraphics[width=0.45\linewidth,trim={0 12.7cm 26.3cm 0},clip]{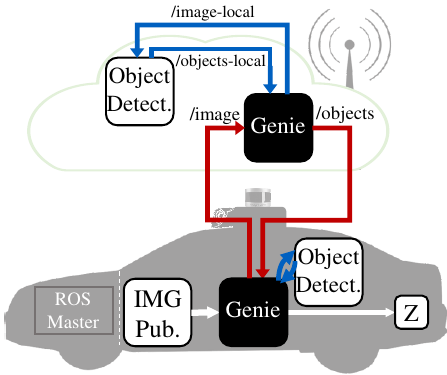}
    \caption{Left: an autonomous car connected to the edge. Right: an example of applying Genie in an autonomous vehicle.}
    \vspace{-3mm}
    \label{fig:1Car-example}
\end{figure}

\noindent\textbf{Proposed Edge Benefits.} Fig.~\ref{fig:1Car-example} shows a scenario where a car is connected to the edge.  As is evident in the figure, the necessary setup for autonomous vehicles is for the vehicle itself to have every service required for full autonomy (depicted as Node X to Node D). This is a requirement because the connection to the edge-based server could be cut off at any given moment. In the scenario of Fig.~\ref{fig:1Car-example}, the network provider has created a duplicate of a service (Node Y) that already exists on the car. The red arrows in the figure indicate that every message coming out of Node X to Node Y is duplicated both on the edge and on the car. This is done by replicating all the subscribed ($T_S(Y)$) and published ($T_P(Y)$) topics of Node Y to the edge. The results are then fed into Node Z. In this configuration, it is possible for Node Z to receive duplicated messages, and has to identify and discard one of the duplicates (since the services are identical). This is the most common scenario where the edge can be useful. For example, Autoware relies on object detection for tracking and sign detection functionality. However, slow object detection is still serviceable without a remote connection. The value of the edge here is to run faster object detection and thus, increase the overall decision-making accuracy of Autoware. The other potentially useful scenario is for the edge to have an extra service that would improve accuracy when available, depicted as Node F in Fig.~\ref{fig:1Car-example}. For example, edge devices can offer a vision assistant for the car (as we shall discuss in Sec.~\ref{sec:eval:case}). In both scenarios, the connection to the edge shall remain optional.

\begin{figure}[t]
    \centering
    \includegraphics[width=0.9\linewidth,trim={0 15cm 17.3cm 0},clip]{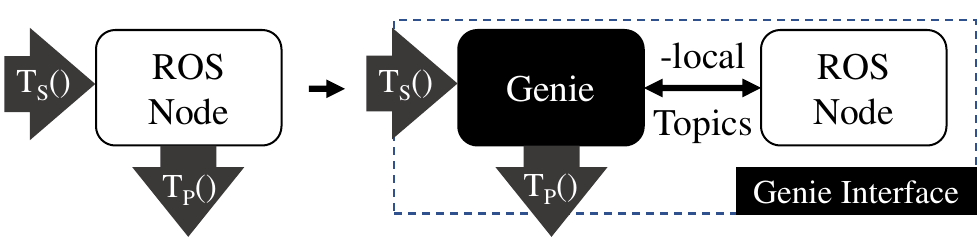}
    \caption{Overview of the Genie.}
    \vspace{-7mm}
    \label{fig:DGC}
\end{figure}



\subsection{Genie ROS Node for Non-Intrusive Caching}\label{sec:des:ros-cache}
With the previously mentioned design goals in mind, the master node inside the vehicle, together with all the service nodes located on the car and the nodes offered by the edge, creates a virtual private network to facilitate peer-to-peer communication between all nodes (which is typical in ROS). In this section, we would like to design a system that can enable efficient caching between these inter-network ROS nodes (we discuss caching across different virtual private networks of ROS in Sec.~\ref{sec:des:dis-cache}).

The main problem to address here is that of building and storing the cache. This problem is not as intuitive as it seems. Imagine Autoware, where ROS nodes are already implemented and deployed. Even though Autoware is open-source, other ROS-based autonomous driving software may not be. Thus, we would like to store the cache in a non-intrusive way without modifying the source code of Autoware's ROS nodes. For that, we present the concept of Genie, an encapsulation made for ROS nodes, presented in the right of Fig.~\ref{fig:1Car-example}. As is evident in the figure, a ROS node is encapsulated in a Genie interface conceptually. This is done by attaching an additional generic Genie ROS node developed by us to each ROS node except the master. This attachment procedure is done by automatically detecting the ROS topics subscribed to and published by that node (depicted as $T_S()$ and $T_P()$ in Fig.~\ref{fig:DGC}), switching them to topics affixed with \texttt{-local} so that they would only send and receive messages to the Genie node, and exposing the original topics by the Genie node itself. To other ROS nodes, this new Genie will act like the encapsulated node. However, before passing on any messages from other ROS nodes, the Genie node can check the local cache. Fig.~\ref{fig:1Car-example} shows this design applied to the Object Detection service of an autonomous vehicle.

\subsection{Distributed Collective Cache}\label{sec:des:dis-cache}

\begin{figure}[t]
    \centering
    \includegraphics[width=0.9\linewidth,trim={0 12.4cm 18.4cm 0},clip]{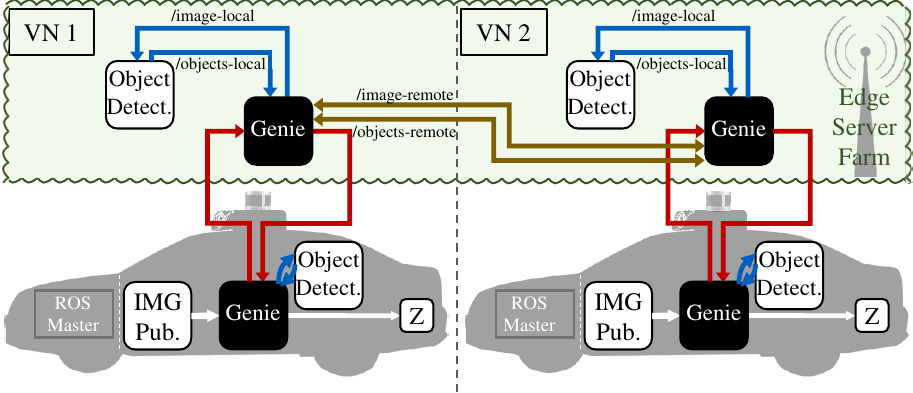}
    \caption{An example of applying Genie to object detection across autonomous vehicles.}
    \label{fig:DG-example-2Car}
    \vspace{-5mm}
\end{figure}

Fig.~\ref{fig:DG-example-2Car} shows a scenario where two cars are connected to the edge. The existence of the ROS virtual network, depicted as VN1 (Virtual Network 1) and VN2, means the "experiences" of the cars would be local to the nodes that are in their respective virtual network. If any cache exists on the edge or on the car, it would only be aware of the information that has been seen by the car itself. For example, with a simple application of Genie, if a vehicle were to be looping around a city block, it could have reduced computation the next time it arrives at the same spot because much of the environment has been seen before. This can be particularly useful for daily commutes, which is the most common use-case for vehicles~\cite{nazari2018shared}. Nonetheless, this situation is limited.

What is more desirable, however, is for the vehicle to receive additional information from other vehicles (i.e., use their experiences). Since each car has its virtual private network by design, this would be an inherently distributed system. Before exploring this type of collective cache construction from distributed entities, we first present a unified interface that is recognized by all members of the system (i.e., topics that all Genie ROS nodes are aware of).

\noindent\textbf{Cache Sharing Interface.} As part of our design goal, we would like to expand the Genie ROS node so that it is capable of communicating with other Genie nodes in a distributed fashion. To achieve that, each Genie will expose an additional set of topics with a \texttt{-remote} label affixed for the topics detected on the encapsulated node. To clarify, imagine the scenario of Fig.~\ref{fig:DG-example-2Car}. As we discussed in Sec.~\ref{sec:des:ros-cache}, the existing ``/image" and ``/objects" topics will be changed to \texttt{``/image-local''} and \texttt{``/objects-local''} and the original topics will be taken over by the Genie node. To inform other Genies that such services exist, the Genie node also publishes messages selectively on the \texttt{``/image-remote"} and \texttt{``objects-remote"} topics. Likewise, the Genie node will subscribe to the same set of topics to receive information from other remote Genie nodes. The simplicity of these interfaces means that if a Genie has encapsulated an object detection node, it can understand and communicate with other Genies that are also object detection nodes, or understand nodes that work with a subset of the same topics. This design would make sharing functionalities across the network less sophisticated.

\noindent\textbf{Collective Cache Construction.} Alg.~\ref{alg:collective-cache} depicts how the cache is populated in detail, using the aforementioned interfaces. Whenever a message arrives on a topic, it is checked against the database (line 12). If the entry does not exist, the message is forwarded both on the \texttt{-local} version (line 13) and the \texttt{-remote} of that topic. If the entry exists in the database (line 16), the stored results are retrieved (line 17) and directly returned to the sender (line 19). 
Sending messages on the \texttt{-remote} topics constitutes broadcasting the message on the edge network if the Genie is on the edge, where other Genies with the same functionality have subscribed to the same set of topics  (whereas sending message on \texttt{-remote} for local car-based Genies uploads the workload to the remote Genie). Upon receiving a message from neighboring Genies, the same procedure of cache miss (line 12) and cache hit (line 16) is followed. If the Genie has an entry for the request, the results will be sent back to the sender directly (line 19). We shall explain line 9 and line 18 in our enhanced cache design, discussed in Sec.~\ref{sec:des:sem-cache}.

\begin{algorithm}[!t]
\setstretch{1.1}
\caption{Genie ROS Node Distributed Cache Procedure}
\renewcommand{\algorithmicensure}{\textbf{Output:}}
\let\oldnl\nl
\newcommand{\nonl}{\renewcommand{\nl}{\let\nl\oldnl}}
\label{alg:collective-cache}
\scriptsize{
\begin{algorithmic}[1]
\Require $M<header, data, T>$ \Comment{The message M with a header, data, topic.}
\Require $T<name, type>$ \Comment{Topic T with a name and a type.}
\Require $H_{DB}$ \Comment{Database for all the hashmaps.}
\Require $PUBLISH(M, T)$ \Comment{Publishes a message M on the topic T.}

\Function{BuildHashMap}{T}
    \State $\text{hashmap}_T = \langle \text{type\_of}(T) , \{ \} \rangle$
    \State $H_{DB}.\text{add}(\text{hashmap}_T)$
\EndFunction
\Function{OnMessageArrival}{M, T}
    \If{$T \notin H_{DB}$}
        \State \Call{BuildHashMap}{T} \Comment{Create a new hashmap for never-seen topics.}
    \EndIf
    \State $\text{cacheEntry} = \text{LookUpByHeader}(\text{hashmap}_T, M.\text{header})$
    \If{$\text{cacheEntry} \neq \text{null}$} \Comment{Message is an answer to our previous query.}
        \State \Call{EnhancedCacheNewData}{M} \Comment{See Algorithm~\ref{alg:object-map}}
        \State $\text{cacheEntry.second} = M.\text{data}$ \Comment{Store it in the hashmap as value.}
        \State \Return
    \ElsIf{\Call{LookUpLocalCache}{$\text{hashmap}_T, M = \text{null}$}} \Comment{\textbf{Cache miss.}}
        \State \Call{PUBLISH}{$M$, $T$ + ``-local''}  \Comment{Share with the encapsulated node.}
        \State \Call{PUBLISH}{$M$, $T$ + ``-remote''} \Comment{Share with remote Genie.}
        \State $\text{hashmap}_T.\text{add}(M, \{ \})$ \Comment{Add message as key with empty value.}
    \Else
        \State $\text{cacheEntry} = \Call{LookUpLocalCache}{\text{hashmap}_T, M}$ \Comment{\textbf{Cache hit.}}
        \State \Call{EnhancedCacheBoostData}{cacheEntry} \Comment{See Algorithm~\ref{alg:object-map}}
        \State \Call{PUBLISH}{$\text{cacheEntry.second}, \text{cacheEntry.second}.T$}
    \EndIf
\EndFunction
\end{algorithmic}
}
\end{algorithm}




\subsection{Driving-specific Caching}\label{sec:des:sem-cache}

For autonomous vehicles, using an object map (i.e., a map where individual records are of specific detected objects such as a traffic light) instead of a generic message-based ROS cache can have several benefits. In this section, we detail a design that is specific to object caching.

\noindent\textbf{Smart Cache Specific to Autonomous Driving.} To facilitate our design, we have identified key data structures required in autonomous driving, which include raw images, LIDAR point clouds, and 3-dimensional objects. Specifically in Autoware, objects include every information needed to enable the computation-heavy perception module, eventually leading to autonomous driving decisions. To reiterate, storing objects instead of raw data such as images has several benefits: First, the objects are already-processed information, and contain more useful data whereas raw data must be processed first. Second, objects have substantially smaller storage and communication overhead because they are orders of magnitude smaller than raw data. Third, an object can be more easily translated into 3-dimensional space. As an example, for two cars that are moving in opposite directions, raw image data cannot be useful because images cannot be easily rotated and translated on their z-axis. Meanwhile, the three-dimensional coordinates of objects can be translated in any direction (location translation is one of the main techniques used in autonomous vehicles already~\cite{fu2015path}). For example, a traffic light with location $<x_1,y_1,z_1>$ relative to car 1 can be translated to $<x'_1,y'_1,z'_1>$ for car 2 based on their ground locations. Thus, we use objects exclusively to build our driving-specific cache in addition to the message-based caching proposed in Sec.~\ref{sec:des:dis-cache}.



\noindent\textbf{Location-based High Confidence Object Map.} As keen readers have noticed, lines 9 and 18 in the previous Alg.~\ref{alg:collective-cache} call upon an enhanced cache before storing and retrieving data. Alg.~\ref{alg:object-map} depicts the implementation for those two aforementioned functions. First and foremost, upon data arrival, the function \texttt{EnhancedCacheNewData} is called. Each object in the received objects array is checked against the $OBJECT\_MAP$ database of type $<location,objects>$ in line 4. We use absolute object location as the key to the object map since absolute location is the only identifying characteristic of an object that multiple cars can agree on. If not in the database, line 7 would add the object.

A key design decision was to decide on how the database treats object confidence. A key feature of machine learning techniques such as DNN used for object detection is that they can calculate a score, indicating for example how confident they are in labeling an object as a tree. This value is usually between 0 and 1 with 1 being 100\% sure. In this paper, we have decided to create a high-confidence map, meaning that the Genie will only share objects that are above a certain threshold with the cars requesting information. This decision was because information retrieved from the edge has to be reliable. To facilitate that, the threshold $C_T$ for the high-confidence object map is set to be the 60th percentile of the confidence range in our design.

The initial confidence of a stored object could be lower than this threshold (for example, 0.3 instead of 0.65). However, if multiple cars see and detect the same object (potentially from various angles), the confidence in that object could be improved over time. For example, a traffic light is stationary. Even with an initial low score, many cars will see and recognize it. Thus, we would update the confidence of a seen-before object (line 5 of Alg.~\ref{alg:object-map}). We use gradient ascent to update the score.

Finally, Genie will call \texttt{EnhancedCacheBoostData} whenever it is returning a set of objects to the sender. This function checks for existing objects at relevant locations, and adds them to the list of objects to be returned to the sender.

\begin{algorithm}[!t]
\setstretch{1.1}
\caption{Enhanced Semantic Cache}
\renewcommand{\algorithmicensure}{\textbf{Output:}}
\label{alg:object-map}
\scriptsize{
\begin{algorithmic}[1]
\Require $TYPES = \{objects\}$ \Comment{Recognized type for the enhanced cache.}
\Require $M<header, data, T>$ \Comment{The message M.}
\Require OBJECT\_MAP $<location , objects>$  \Comment{Object map.}
\Require $C_T$ \Comment{Confidence threshold for the database (between 0 and 1).}
\Function{EnhancedCacheNewData}{M}
    \If{type\_of(M) $\in TYPES$} 
    \For{object $\in$ M.data.objects}
        \If{$O = OBJECT\_MAP[object.location]$ exists}
            \State $O.C = O.C_l + \lambda \times (O.C_l - M.C_l$) 
        \Else
                    \State OBJECT\_MAP[M.data.location].add(object)
        \EndIf
            \EndFor
    \EndIf
\EndFunction

\Function{EnhancedCacheBoostData}{cacheEntry}
    \If{type\_of(M) $\in$ TYPES} 
        \For{object $\in$ cacheEntry.value.objects}
            \For{stored\_object $\in$ OBJECT\_MAP[object.location]} 
                \If{$\text{stored\_object}.C_l \geq C_T$} 
                \State cacheEntry.value.objects.add(stored\_object) 
                \EndIf
            \EndFor
        \EndFor
    \EndIf

\EndFunction
\end{algorithmic}
}
\end{algorithm}

\section{Evaluation}\label{sec:eval}

\subsection{Experimental Setup}

\noindent\textbf{Devices.} We evaluate our design's efficacy in tail latency, reusability, and value-added accuracy using an edge server cluster modeled after our industry partner's platform. The cluster includes three NVIDIA AGX Xaviers and three NVIDIA Jetson TX2s. The Jetson TX2s represent cars running the full autonomous driving software plus the ROS master, while the AGX Xaviers serve as remote edge servers. For heterogeneous setups, we add a server with an Intel Xeon E5-2650 CPU and NVIDIA Quadro RTX 4000.

\noindent\textbf{Configuration.} Cars and edge servers are co-located to eliminate network delays, focusing solely on computational latency effects from our caching method. We use Autoware, an open-source autonomous driving software gaining industry traction~\cite{wiggers_2019}. Other ROS-based alternatives like those from BMW and Baidu share similar architectures~\cite{aeberhard2015automated,zhang2017powering}.

\noindent\textbf{Data.} The KITTI Vision Benchmark suite~\cite{geiger2015kitti} provides raw data for playback from vehicle sensors, including four camera streams, a LIDAR point cloud, and precise location.

\noindent\textbf{Measurements.} Metrics include tail latency, image reusability (cache hits/total cache requests for image cache), object reusability (cache hits/total cache requests for object map), and confidence boost (average total boost for object map).

\noindent\textbf{Scenarios.} We evaluate scenarios with 1, 2, and 3 cars to explore different levels of complexity. Increased car involvement enhances collective cache information.

\noindent\textbf{Homogeneous vs. Heterogeneous Cluster.} We test clusters with low-power devices and add a powerful server for heterogeneous configurations. The Quadro-based server acts as one of the remote machines in heterogeneous tests.

\noindent\textbf{Genie Case Study.} We perform a case study demonstrating how Genie aids vision-assisted cars through communication with vision-enabled remote Genies.

\subsection{Tail-Latency Performance}

\begin{figure}[t]
\centering

\hspace*{10mm}\includegraphics[width=0.45\linewidth]{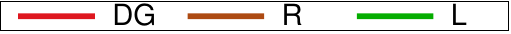}\vspace{-3mm}
\resizebox{0.495\textwidth}{!}{
\subfloat[][1 car.]{\includegraphics[width=0.18\textwidth]{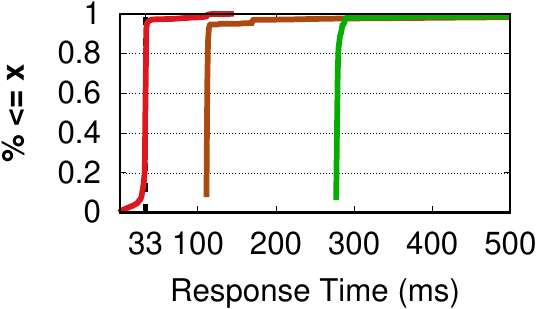}\label{fig:1-1Car}}
\subfloat[][2 cars.]{\includegraphics[width=0.162\textwidth, trim= 0.9cm 0 0 0, clip]{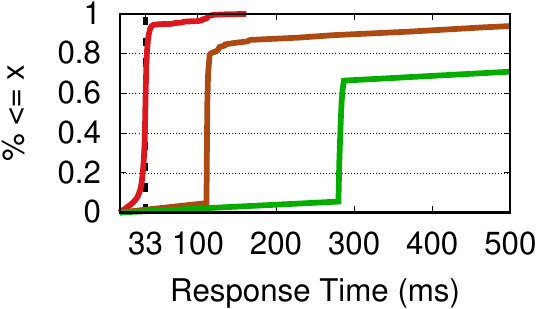}\label{fig:1-2Car}}
\subfloat[][3 cars.]{\includegraphics[width=0.162\textwidth, trim= 0.9cm 0 0 0, clip]{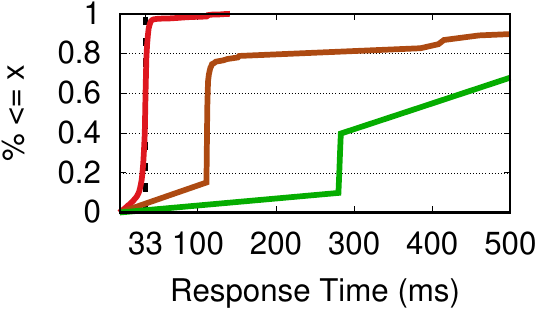}\label{fig:1-3Car}}
}

(i) Homogeneous setting.

\vspace{2mm}
\hspace*{10mm}\includegraphics[width=0.45\linewidth]{figs/key-time.pdf}\vspace{-3mm}
\resizebox{0.495\textwidth}{!}{
\subfloat[][1 car.]{\includegraphics[width=0.18\textwidth]{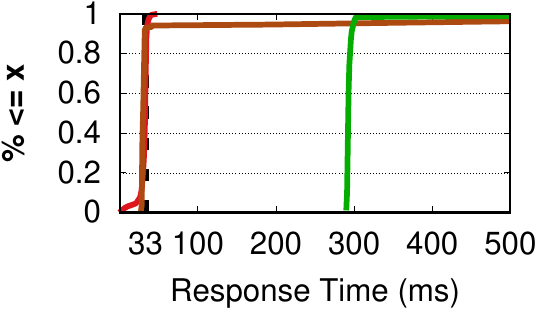}\label{fig:2-1Car}}
\subfloat[][2 cars.]{\includegraphics[width=0.162\textwidth, trim= 0.9cm 0 0 0, clip]{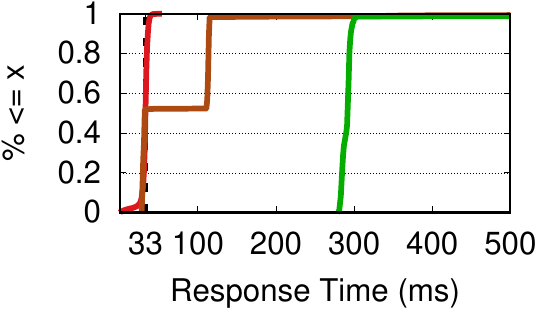}\label{fig:2-2Car}}
\subfloat[][3 cars.]{\includegraphics[width=0.162\textwidth, trim= 0.9cm 0 0 0, clip]{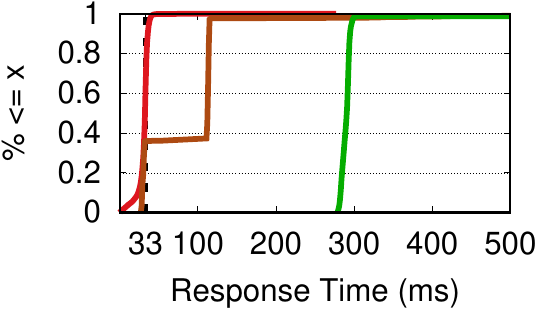}\label{fig:2-3Car}}
}

(ii) Heterogeneous setting.
    \caption{Cumulative distribution function for the response time of Genie(DG), versus remote (R) and local (L) ROS execution on homogeneous and heterogeneous configurations.}
    \label{fig:cdf-time}
    \vspace{-5mm}
\end{figure}

Fig.~\ref{fig:cdf-time} shows the cumulative distribution function (CDF) of response times for Genie (DG) versus ROS remote (R) and local (L) execution of the Autoware Vision Detector (AVD) across scenarios with 1, 2, and 3 cars in homogeneous and heterogeneous edge clusters. The 33ms deadline is marked with a dashed line. Genie outperforms both remote and local executions of AVD, achieving an average improvement of 82\%, consistently meeting real-time demands with stable response times due to efficient caching and selective execution of algorithms. Adding more cars does not significantly impact latency, even with network communication, due to the wired Ethernet setup.

In the 1-car heterogeneous scenario, the server (R) performs similarly to Genie initially, but as cars increase, remote execution experiences higher latency, showcasing Genie’s advantage in maintaining consistent performance. Genie’s execution time remains stable across low-power embedded and powerful heterogeneous servers, thanks to its efficient CPU operations, adaptable to both ARM and Intel CPUs~\cite{larabel}. This ensures Genie’s reliability across various architectures.

\subsection{Image Caching}
\begin{figure}[t]
\centering
\hspace*{10mm}\includegraphics[width=0.65\linewidth]{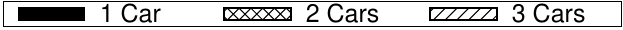}\vspace{-2mm}
\subfloat[][Homogeneous setting.]{
    \includegraphics[width=0.5\linewidth]{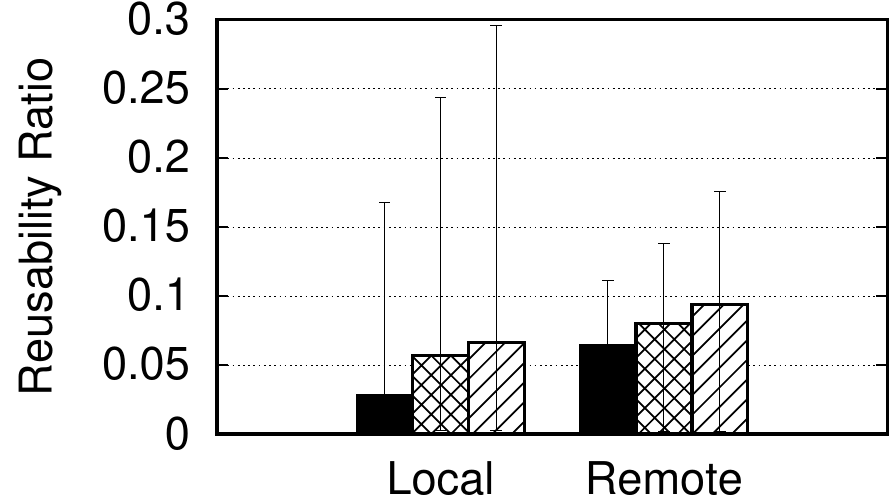}\label{fig:imgrr-reuse-ho}}
\subfloat[][Hetero. setting.]{
    \includegraphics[width=0.2553\linewidth]{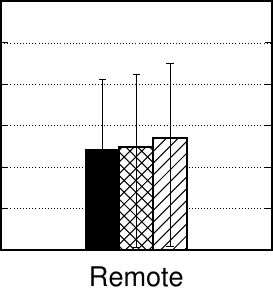}\label{fig:imgrr-reuse-ht}}
    
 \caption{Image reusability ratio for local Genies and remote Genies under the three scenarios.}
    \label{fig:reuse-img}
    \vspace{-4mm}
\end{figure}

In this section, we measure the reusability of the average image cache under DG for the three scenarios and both on the homogeneous and heterogeneous server clusters. The results are depicted in Fig.~\ref{fig:reuse-img}. The results are divided into Local, where the data is recorded on the cars (i.e., TX2) versus Remote, where the data is recorded from the remote servers (i.e., AGX). As is evident in the figure, Remote Genie nodes are capable of reusing more data because they can communicate with other cars in the case of 2-car and 3-car scenarios. In the case of the 1 Car scenario, the remote Genie has faster hardware and thus can collect many more objects compared to the local Genie. Fig.~\ref{fig:reuse-img} also shows the error bars. As is evident in the figure, the maximum reusability on the local cache for the 2 cars and 3 cars scenario is quite high because the cache can be inflated initially from the indirect communication with the remote Genie of other cars.

Finally, the faster server in the heterogeneous architecture has enabled the collective cache to gather processed images faster, leading to higher average and maximum reusability.

\begin{figure}[t]
\centering
\hspace*{8mm}\includegraphics[width=0.45\linewidth]{figs/key.pdf}\vspace{-2mm}
\subfloat[][Homogeneous setting.]{
    \includegraphics[width=0.5\linewidth]{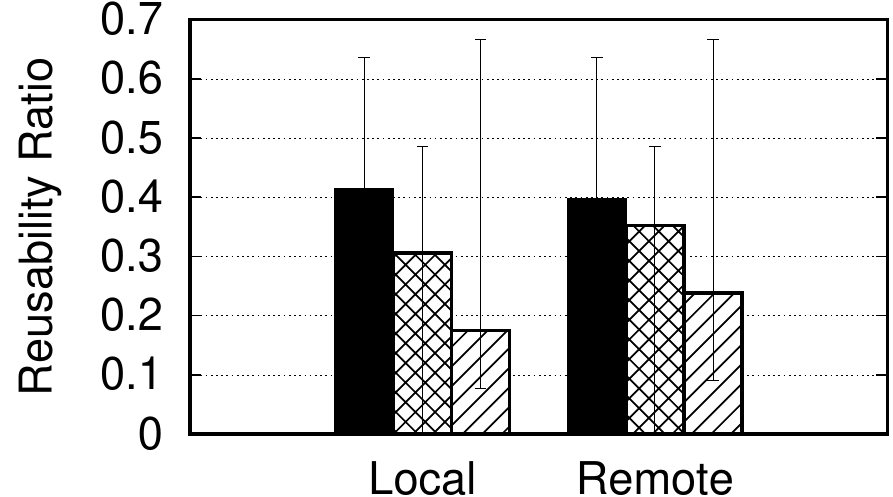}\label{fig:obrr-reuse-ho}}
\subfloat[][Hetero. setting.]{
    \includegraphics[width=0.2553\linewidth]{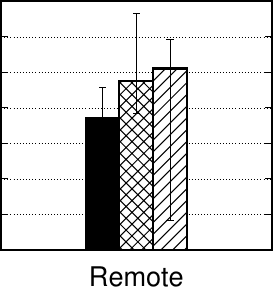}\label{fig:objrr-reuse-ht}}
 \caption{Object reusability ratio for local Genies and remote Genies under the three scenarios.}
    \label{fig:reuse-obj}
    \vspace{-6mm}
\end{figure}

\subsection{Object Caching}

As discussed in Sec.~\ref{sec:des:sem-cache}, object mapping is particularly beneficial for autonomous driving, despite being less general and scalable than our message-based cache. Fig.~\ref{fig:reuse-obj} illustrates the average object reusability, defined as the ratio of cache hits on objects to total requests. The reusability rate for object caching is significantly higher than for image caching, averaging 31\% and reaching a maximum of 67\%, which is four times greater than image cache rates. In homogeneous scenarios, object reusability decreases with more cars because each location can contain a large number of objects. While more cars increase the number of available objects, not all meet the high confidence threshold from Sec.~\ref{sec:des:sem-cache}, slightly reducing overall reusability but enhancing data quality. For the 1-car scenario, the local object map shows better reusability due to selective uploads that avoid unnecessary requests to the remote server when objects are cached locally. In heterogeneous scenarios, the powerful server significantly enhances object reusability, especially with more cars. The server can process four times as many images (and thus objects) compared to the Jetson AGX Xavier, overcoming the limitations seen in the homogeneous setup and leading to increased reusability with additional vehicles.

\subsection{Confidence Boost}
\begin{figure}[t]
    \centering
    \hspace*{5mm}\includegraphics[width=0.7\linewidth]{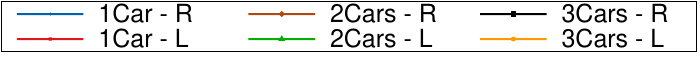}\vspace{-2mm}
    
    \subfloat[Homogeneous setting.]{
        \includegraphics[width=0.45\linewidth, trim = 0 0 0 1.4cm, clip]{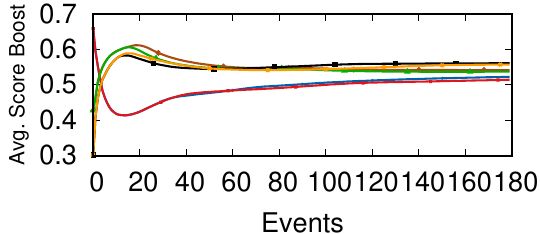}
        \label{fig:boost-ho}
    }
    \hfill 
    \subfloat[Heterogeneous setting.]{
        \includegraphics[width=0.45\linewidth, trim = 0 0 0 1.4cm, clip]{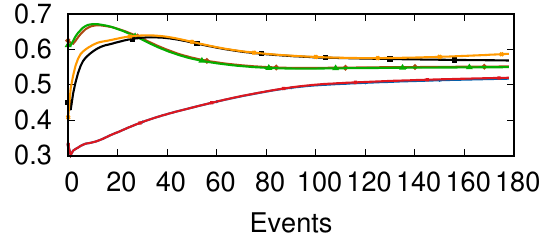}
        \label{fig:boost-ht}
    }

    \caption{Cumulative average for the score (confidence) boost of all objects in remote and local Genies for all scenarios.}
    \vspace{-4mm}
    \label{fig:boost}
\end{figure}

We measured the confidence boost of our method across various scenarios, as shown in Fig.~\ref{fig:boost}. The cumulative average scores results are compared for 1, 2, and 3 cars in both homogeneous and heterogeneous architectures. Confidence boosts are recorded for both remote (R) and local (L) Genies. The y-axis shows the running average confidence addition per object, while the x-axis represents the number of recorded events. Local Genies have generally lower confidence due to limited communication. The 3-car scenario shows a higher confidence boost compared to 2-car and 1-car scenarios, and the heterogeneous architecture achieves a higher cumulative average due to faster processing by the powerful server. Over a long period, the more cars that are present in the system, the higher the quality of the data on the remote Genies will become. Moreover, the heterogeneous architecture can achieve a slightly higher cumulative average value due to the faster processing of the powerful server.

\noindent\textbf{Benefits of Distributed Cache versus Local Cache.}
Fig.~\ref{fig:reuse-obj} and Fig.~\ref{fig:boost} demonstrate the superiority of distributed caching in enhancing information sharing and reusability among nodes, a feature unattainable with local caches in ROS. Additionally, our distributed cache maintains minimal overhead, averaging 8.8ms (for a mix of cache hits and cache misses).


\subsection{Case Study: Vision-Assisted Driving}\label{sec:eval:case}
\begin{figure}
    \centering
    \includegraphics[width=0.55\linewidth]{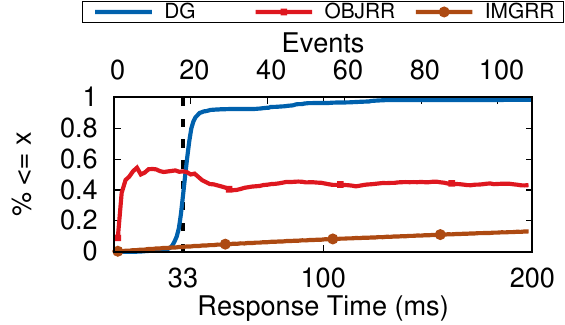}
    \caption{Vision-assisted driving: the CDF for time (ms), object reuse rate (OBJRR) based on events for the second car with no object detection (the image reuse rate (IMGRR) from the first car is also shown for reference).}
    \label{fig:bs}
    \vspace{-3mm}
\end{figure}

In our case study on vision-assisted driving, we explored the potential of enhancing autonomous vehicles that navigate without cameras, relying instead on LIDAR and high-definition maps for localization, and point cloud-based detectors and radars for obstacle avoidance~\cite{lachachi2019trueview}. We introduced phantom Genies in our Genie design to assess if data from camera-equipped vehicles could benefit these non-vision-based systems. Our coordinator server identifies vehicles lacking Autoware Vision Detector and assigns them local and remote phantom Genies, facilitating data sharing among vehicles. For instance, in a two-car setup, one vehicle can utilize data from the other, which is illustrated in our findings. As shown in Fig.~\ref{fig:bs}, although the response time for a car using this shared information is slightly increased, it remains under 41ms or 24FPS, demonstrating the viability of this approach despite the communication delay. 

\section{Related Work}\label{sec:relatedwork}

\noindent\textbf{Caching and Intelligent Caching.} The concept of reusing computation to minimize costs is well-established. Techniques like FoggyCache enable close-quarters devices to reuse redundant computations~\cite{guo2018foggycache,cashier}, but depend on centralized servers and lack driving-specific benefits. Similar methods exist for cloud environments~\cite{crankshaw2017clipper,MCDNN}, relying on centralized APIs~\cite{abadi2016tensorflow,cui2016geeps} to manage caching, applicable to AR/VR~\cite{slocum2023going}, storage systems~\cite{storageCache}, mobile apps~\cite{mach2017mobile,Drolia2017PrecogPF}, and robotics~\cite{li2023mimonet}. A few concurrent works like FogROS~\cite{chen2021fogros}, FogROS2~\cite{ichnowski2023fogros}, and Schafhalter et. al \cite{schafhalter2023leveraging} also focus improving latency performance in robots by using cloud hardware.

\noindent\textbf{AI-based Collaborative Sensing.}
AI is expected to make great impacts on various robotic fields, including home services~\cite{you2003development}, healthcare~\cite{ma2023learning,ma2022learning,lyu2022multimodal,lyu2024badclm}, and transportation~\cite{ma2023eliminating,ma2024data,cottam2024large,zhang2024large}, etc., in 2030 according to Stanford's report~\cite{stone2022artificial}. Deep learning approaches have become prominent in autonomous driving~\cite{dong2023deep,dong2023graph}. Approaches like collaborative sensing~\cite{lu2019collaborative,grulich2018collaborative} focus on individual devices or central units for task distribution, as seen in Potluck~\cite{potluck} and Darwin phones~\cite{darwin}. This contrasts with our decentralized approach. EMP~\cite{zhang2021emp} and AutoCast~\cite{qiu2021autocast} proposed collaborative perception leveraging the edge to improve overall accuracy via decentralized data
sharing among vehicles. Genie has the potential to integrate these solutions in ROS-based scenarios.

\noindent\textbf{Real-Time Autonomous Driving.} Reusability in distributed systems is uncommon in real-time contexts. Caching offers significant potential for meeting latency requirements, as seen in embedded platforms using approximation~\cite{apnet}, scheduling~\cite{gpusync,li2023red,LI2021101936,li2023rt}, memory management~\cite{rodolfo1,DBLP:conf/rtss/0001SLS0L23}, and software-hardware co-design~\cite{zhang2021hardware}. Our solution Genie focuses on autonomous vehicle caching and introduces unique challenges, such as non-intrusive cache management and collaborative caching. Furthermore, we plan to adapt Genie to more safety-critical multi-robot scenarios~\cite{zhang2020manipulator,gao2023autonomous,gao2024decentralized,gao2024adaptive,yu2024stochastic}.

\section{Conclusion}

This paper addresses three key challenges identified by an industry partner that impede the practical implementation of edge computing for connected autonomous vehicles. We developed Genie, demonstrating its effectiveness through implementation and evaluation with realistic workloads and contemporary platforms. Future work aims to extend Genie beyond autonomous driving and adapt it to ROS 2 for more modern robotic system solutions.

\clearpage

\hypersetup{breaklinks=true}
\urlstyle{same}
\Urlmuskip=0mu plus 1mu
\bibliographystyle{IEEEtran}
\bibliography{references}

\end{document}